\documentclass[runningheads]{llncs}
\usepackage[T1]{fontenc}
\usepackage{booktabs}       
\usepackage{amsfonts}       
\usepackage{nicefrac}       
\usepackage{microtype}      
\usepackage{xcolor}         
\usepackage{makecell}
\usepackage[most]{tcolorbox}
\usepackage{algorithm}
\usepackage{algorithmic}
\usepackage{amssymb}
\usepackage{amsmath}
\usepackage{bm}
\usepackage{dsfont}
\usepackage{multirow}
\usepackage{color}
\usepackage{subcaption}
\usepackage{booktabs}

\usepackage{makecell}

\usepackage{diagbox}
\usepackage{minitoc}
\usepackage{graphicx}
\usepackage[table]{xcolor}
\usepackage{wrapfig}
\usepackage{bbding}

\begin{document}
\title{Selective Forgetting for Large Reasoning Models}
%
%
\author{Tuan Le \and
Wei Qian \and
Mengdi Huai}
\authorrunning{T. Le et al.}
%
\institute{
Iowa State University, Ames IA 50011, USA \\
\email{\{tuanle,wqi,mdhuai\}@iastate.edu}}

\maketitle            

\begin{abstract}
Large Reasoning Models (LRMs) generate structured chains of thought (CoTs) before producing final answers, making them especially vulnerable to knowledge leakage through intermediate reasoning steps. Yet, the memorization of sensitive information in the training data such as copyrighted and private content has led to ethical and legal concerns. To address these issues, selective forgetting (also known as machine unlearning) has emerged as a potential remedy for LRMs. However, existing unlearning methods primarily target final answers and may degrade the overall reasoning ability of LRMs after forgetting. Additionally, directly applying unlearning on the entire CoTs could degrade the general reasoning capabilities. The key challenge for LRM unlearning lies in achieving precise unlearning of targeted knowledge while preserving the integrity of general reasoning capabilities. To bridge this gap, we in this paper propose a novel LRM unlearning framework that selectively removes sensitive reasoning components while preserving general reasoning capabilities. Our approach leverages multiple LLMs with retrieval-augmented generation (RAG) to analyze CoT traces, identify forget-relevant segments, and replace them with benign placeholders that maintain logical structure. We also introduce a new feature replacement unlearning loss for LRMs, which can simultaneously suppress the probability of generating forgotten content while reinforcing structurally valid replacements. Extensive experiments on both synthetic and medical datasets verify the desired properties of our proposed method.

\end{abstract}
\section{Introduction}
\label{sec:intro}

Under the data-driven paradigm, Large Reasoning Models (LRMs) have demonstrated promising performance and showcased immense potential in further promotion. Unlike standard Large Language Models (LLMs) that directly produce final answers, LRMs are designed to generate structured intermediate reasoning traces before producing their final answers. These reasoning traces, often referred to as chains of thought CoTs, enable LRMs to exhibit multi-step logical inference and compositional reasoning across diverse domains. Recently, a growing number of state-of-the-art LRMs have been proposed~\cite{jaech2024openai,guo2025deepseek}, achieving broad adoption in both research and industry due to their enhanced reasoning ability.

In practice, ensuring responsible usage of LLM services is of utmost importance, given the lessons of previous privacy violations \cite{jeon2023use}. As LRMs are trained on large-scale web and user-generated data, they can inadvertently internalize copyrighted, private, or otherwise sensitive information within their reasoning processes~\cite{jiang2025safechain,carlini2021extracting,wu2025effectively}. This raises significant pressing ethical and technical questions. To ensure data safety, data privacy regulations such as the General Data Protection Regulation (GDPR)~\cite{regulation2018general} grant individuals \emph{the right to be forgotten}, i.e., the ability to revoke the use of their data by models. However, retraining large models from scratch to remove such data is computationally prohibitive and often infeasible due to the inaccessibility of the original pretraining data. Hence, a new paradigm of selective forgetting (also known as machine unlearning)~\cite{qian2022patient,liu2022continual,qian2023towards,zhao2023static,yao2024large,chen2025survey} has emerged, which aims to efficiently remove the influence of specific data from pretrained models without requiring full retraining.

Recently, many unlearning methods have been proposed for LLMs, such as Gradient Ascent (GA)~\cite{yao2024large}, Gradient Difference (GD)~\cite{liu2022continual}, and Preference Optimization (PO)~\cite{maini2024tofu}. However, these unlearning methods are largely designed for models without explicit reasoning structure and fail to account for the unique reasoning pathways within LRMs. Unlike LLMs, LRMs can embed forgettable information not only in their final answers but also within intermediate reasoning steps. Since reasoning traces may encode sensitive, private, or memorized content from training data, it is imperative to assess whether LRMs can be induced to forget such information when required. Therefore, unlearning only the final outputs is insufficient to prevent leakage of forgotten information through reasoning traces within large reasoning models~\cite{yoon2025r}.

In this paper, we aim to design a novel unlearning method tailed for LRMs. While the prior work \cite{wang2025rethinking} proposes a reasoning-aware representation misdirection unlearning method (R$^2$MU) that maps the internal representations of reasoning traces from the forget set to random vectors to suppress sensitive reasoning, it still suffers from notable limitations. To retain general reasoning ability, \cite{wang2025rethinking} employs auxiliary reasoning datasets for regularizing the unlearning; however, such datasets are often difficult to obtain in practice. Additionally, it applies unlearning indiscriminately to the entire chain of thought (CoT) and final answers, which may degrade the model’s overall general reasoning capability. Specifically, unlearning the entire CoT can inadvertently erase general and transferable reasoning steps, thereby harming the broader reasoning performance of LRMs. The key challenge is how to precisely identify and remove only the sensitive reasoning knowledge while preserving the structural coherence and general reasoning abilities.

To address the above challenges, we propose a novel feature replacement-aware unlearning framework (FRUL) for LRMs, which selectively removes sensitive reasoning components while preserving overall reasoning ability without using another dataset to regularize the unlearning. Our approach operates in two stages. Specifically, in the first step, we propose to analyze each reasoning trace and isolate forget-relevant content, while simultaneously generating structurally consistent replacements that preserve the reasoning flow. More specifically, we leverage a retrieval-augmented generation (RAG) and large language models to perform such analysis. In the second step, rather than applying unlearning to entire CoTs, we introduce a new feature replacement unlearning loss, which explicitly suppresses the probability of generating forget-relevant data while reinforcing the generated structurally consistent replacements. we also incorporate a gradient descent–based loss term to ensure that our framework precisely removes sensitive knowledge while preserving the model’s reasoning ability across the retain set. Our proposed LRM unlearning method selectively bypasses and replaces sensitive reasoning traces, achieving precise unlearning while preserving coherent and reliable reasoning performance. Extensive experiments demonstrate that our proposed LRM unlearning method effectively removes targeted knowledge while preserving the model’s general reasoning performance.

\section{Related Work}
\label{sec:RelatedWork}

The growing recognition of data privacy and safety risks in LLMs has led to increasing interest in LLM unlearning, a promising paradigm that removes the influence of undesirable data without full retraining~\cite{yoon2025r,qian2025towards}. This capability enables a wide range of applications, including the protection of copyrighted and personally identifiable information. However, existing LLM unlearning methods fail to address the unique challenges posed by LRM unlearning, which requires going beyond final answers to explicitly remove sensitive information embedded within intermediate reasoning steps. Although \cite{wang2025rethinking} explores unlearning in LRMs, it suffers from a key limitation: it often produces semantically meaningless reasoning traces for forgotten examples, as the model tends to emit random tokens instead of maintaining structured reasoning. In contrast, our method not only effectively suppresses sensitive reasoning content but also generates logically consistent and structured placeholder reasoning, thereby preserving the model’s ability to produce coherent and meaningful outputs even after unlearning.

Retrieval-augmented generation (RAG) mitigates LLM hallucinations by grounding responses in external knowledge sources. During inference, RAG retrieves relevant documents and provides them as context for the generative model~\cite{lewis2020retrieval}, thereby reducing reliance on internal memory and improving factual consistency. It has also been widely adopted in different applications. For example, \cite{wang2025machine} proposes a lightweight RAG-based framework that simulates forgetting by modifying the retrieval corpus instead of retraining the LLM. By removing or replacing disallowed information in the external repository, the model naturally avoids generating responses based on the forgotten content. However, simply using RAG as a filter over model outputs does not ensure a truly safe LLM, as it only prevents sensitive information from being retrieved. In contrast, we propose leveraging RAG to identify forgettable information and actively remove its internal representations within the model itself through our unlearning method.

\section{Methodology}

Let $D$ denote the full training dataset of examples $(q, c, a)$, where $q$ is a question, $c$ is the CoT reasoning trace, and $a$ is the final answer. Based on the given training dataset $D$, we can train a large reasoning model $M_{\text{original}}$, which is parameterized by the model parameters $\theta_{\text{original}}$. We use $D_f \subset D$ to denote the targeted \emph{forget set}, which contains examples associated with sensitive or undesired knowledge that must be removed. Note that the goal of machine unlearning is to produce a new model $M_{\text{unlearn}}$  that behaves as if $D_f$ had never been included during training, while preserving accuracy on the \emph{retain set} $D_r = D \setminus D_f$. For LRMs, this ensures that selective forgetting precisely unlearns the requested sensitive information while preserving the model’s general reasoning capabilities. 

Note that machine unlearning adapts the model parameters to suppress the probability of generating forgotten outputs while retaining performance on the retaining data $D_r$. However, existing approaches \cite{yao2024large,liu2022continual,maini2024tofu,wang2025rethinking} typically operate on the final answers alone, which is insufficient for LRMs, as sensitive knowledge can also appear within intermediate reasoning steps, potentially leaking forgotten information even if the final prediction appears unaltered. To precisely unlearn the requested target forget information, we propose to resort to a reasoning-aware unlearning objective that targets both intermediate CoTs and final answers, ensuring selective forgetting with minimal degradation of reasoning ability.

For each example $(q, c, a)$, let $c_f \subset c$ denote the subset of reasoning features that directly correspond to forget-relevant information in $D_f$, and let $c_m$ represent a modified reasoning trace where $c_f$ is replaced by benign or randomized placeholders while preserving the overall logical structure of $c$. Our proposed \emph{FRUL} operates explicitly on $(c_f, c_m)$ pairs. The goal is to simultaneously suppress the model’s probability of producing the sensitive reasoning $c_f$ while reinforcing the benign reasoning $c_m$, thereby teaching the model to ``route around’’ forgotten knowledge. For the requested unlearning data $D_{f}$, to degrade model performance on the forget set $D_f$ while preserving accuracy on the retain set $D_r$, we adopt the following formulation based on the gradient difference loss
\vspace{-0.02in}
\begin{align}
\label{eq:loss_gd}
& \ell_{\mathrm{GD}}(\theta; D_f, D_r)
= -\mathbb{E}_{(q,a)\sim D_f}[-\log p_\theta(a \mid q)]\\
& \qquad\qquad\qquad\qquad\qquad + \alpha\,\mathbb{E}_{(q,a)\sim D_r}[-\log p_\theta(a \mid q)].
\notag
\end{align}
\vskip -2pt
\noindent The above loss considers a setting where unlearning is restricted to the final answers associated with samples in the requested unlearning data $D_f$, while maintaining the accuracy of the final answers on the retaining data $D_r$.

However, as discussed earlier, for LRMs, restricting unlearning to the final answers is inadequate, since sensitive information from $D_f$ can still be exposed through intermediate reasoning traces. To address this limitation, we propose a novel selective forgetting objective tailored to LRMs that simultaneously discourages the generation of forget set content in both reasoning steps and final answers, while maintaining general overall reasoning capabilities. The challenge here is how to precisely unlearn the targeted knowledge, without hurting other knowledge. To address this challenge, we propose to perform the feature-level guided knowledge unlearning. To achieve this, we will leverage multiple large language models with RAG to analyze each reasoning example in the forget set $D_f$. Recall that $D_f = {(q, c, a)}$ contains question $q$, reasoning $c$, and answer $a$ for instances we wish to unlearn. The LLMs has access to a knowledge base populated with $D_f$, allowing it to recognize and isolate only the forgettable content $c_f \subset c$ in CoTs, and the remaining reasoning. Specifically, we prompt the large language models by providing a CoT trace alongside a knowledge context retrieved from $D_f$. The LLMs are instructed to extract only those CoT segments that explicitly match or logically derive from the provided forget-relevant knowledge that lead to the answer. Irrelevant reasoning or general logic is excluded, ensuring that only sensitive content is isolated as $c_f$. To improve the reliability of this extraction process, we use distinct large language models to independently extract candidate $c_f$ segments and then aggregate their generated outputs into a unified list to determine the final forgettable content.

Based on the extracted CoT segments, we proceed to perform feature replacement. Specifically, the LLM substitutes each identified segment in $c_f$ with dummy information---benign placeholder content that does not carry the original factual meaning. Importantly, this substitution is done in a way that preserves the logical structure and sequence of reasoning in $c$. We prompt the LLM with the original CoT and the identified forgettable segments $c_f$. The large language model is instructed to rewrite the CoT by substituting all sensitive content with neutral placeholders (e.g., variables or generic terms) while preserving the original logical structure, reasoning flow, and mathematical validity. This ensures the output remains coherent and faithful to the reasoning logic, without disclosing sensitive information. The result is a modified chain-of-thought $c_m$, which follows the same reasoning steps as the original $c$ but with the sensitive knowledge replaced by neutral or unreal facts. Let $p_\theta(c \mid q)$ denote the probability assigned by the model with parameters $\theta$ to generate the sequence $c$ given the input $q$. Based on the above, to simultaneously suppress $c_f$ and reinforce $c_m$, we design the following unlearning objective over the CoT
\vspace{-0.01 in}
\begin{align}
\label{eq:loss_cot}
    & \ell_{\mathrm{CoT}}(\theta, D_f) \;=\; 
\lambda_f \,\mathbb{E}_{(q,c_f)\sim D_f}[-\log(1 - p_\theta(c_f \mid q))]
\; \\
& \qquad\qquad\qquad\qquad\qquad +\;
\lambda_r \,\mathbb{E}_{(q,c_m)\sim D_f}[-\log p_\theta(c_m \mid q)], \notag 
\end{align}

\noindent where $\lambda_f,\lambda_r \geq 0$ are the trade-off hyperparameters. The first term discourages the model from producing the forgotten reasoning $c_f$, while the second term encourages the model to generate the replacement reasoning $c_m$. Through this process, the model learns to prevent the reproduction of forgotten knowledge while preserving coherent and structurally consistent reasoning traces.

In addition to performing unlearning at the chain of thought level, it is also important to preserve the model’s reasoning ability on the retaining data $D_r$ while intentionally degrading its performance on the forget set $D_f$. Without such preservation, the unlearning process could inadvertently impair the model’s general reasoning competence, leading to collateral forgetting across unrelated tasks. To address this, we introduce the following reasoning preservation loss $\ell_{\mathrm{RP}}$ to regularize the model on the retain set $D_r$
\vspace{-0.01 in}
\begin{align}
    \ell_{\mathrm{RP}}(\theta, D_r)
    = \mathbb{E}_{(q,c)\sim D_r}[-\log p_\theta(c \mid q)],
\end{align}

\noindent where $D_r$ is the retaining data. The above objective encourages the model to faithfully reproduce valid and coherent reasoning traces for non-forgotten samples, thereby maintaining its overall reasoning fluency and stability.

To summarize, in order to jointly enforce selective forgetting of sensitive reasoning traces and preservation of general reasoning capability of the resulting unlearned models, the overall loss that we are minimizing is
\vspace{-0.01 in}
\begin{align}
\label{eq:overall_loss}
    \min_{\theta}\; \ell_{\mathrm{FRUL}}(\theta, D_f,D_r) =
    \ell_{\mathrm{CoT}}(\theta, D_f)
    + \beta_{g}\, \ell_{\mathrm{GD}}(\theta; D_f, D_r)
    + \beta_r\,\ell_{\mathrm{RP}}(\theta, D_r),
\end{align}

\noindent where $\beta_r$ controls the strength of reasoning preservation on $D_r$, and $\beta_g$ controls the relative weight of GD loss component. By leveraging multiple LLMs with RAG to isolate forget-relevant reasoning segments and constructing structurally consistent replacements, our method ensures that the model forgets sensitive knowledge without erasing its broader reasoning capacity. Our proposed method enables selective suppression of forgotten content while preserving utility on retained data. This combination of reasoning-aware feature extraction, replacement-based anchoring, and tailored loss design constitutes a novel and principled framework for unlearning in large reasoning models. 

To solve the proposed loss in Eq. (\ref{eq:overall_loss}) on LRMs, we adopt the gradient-based optimization using the AdamW optimizer~\cite{loshchilov2017decoupled}. All components of the overall objective are differentiable with respect to model parameters, allowing standard backpropagation for end-to-end optimization. During optimization, we alternate mini-batches from the forget set \(D_f\) and retain set \(D_r\) to balance the forgetting and retention objectives. A learning rate scheduler with a warm-up phase is also applied to promote smooth convergence. This optimization strategy enables effective unlearning while maintaining the model’s overall reasoning capabilities.

\section{Experiments}
\label{sec:Experiments}

\subsection{Experimental Setup}
\vspace{-0.02 in}

\begin{figure*}[t!]
\centering
\begin{subfigure}{0.495\linewidth}
\includegraphics[width=1\linewidth]{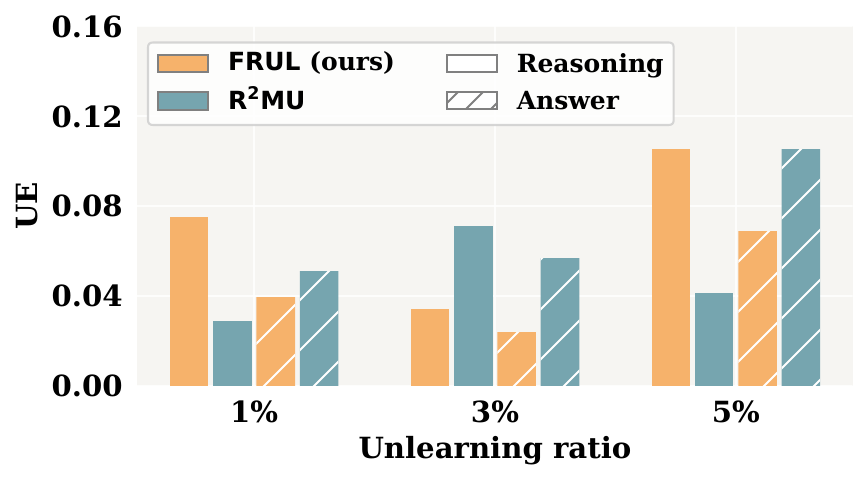}
\vskip -8pt
\caption{Reasoning-Llama-3.2-1B}
\label{fig:unlearn_rtofu_forget_a}
\end{subfigure}
\begin{subfigure}{0.495\linewidth}
\includegraphics[width=1\linewidth]{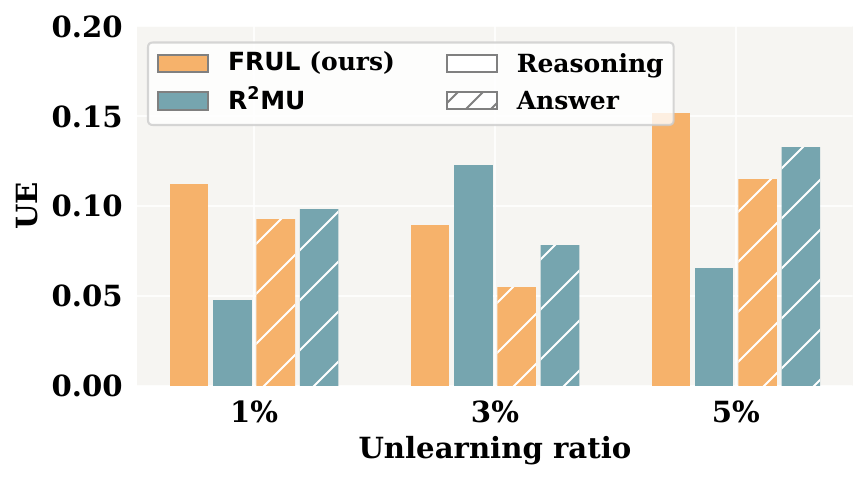}
\vskip -8pt
\caption{Nemotron-Nano-8B}
\label{fig:unlearn_rtofu_forget_b}
\end{subfigure}
\vskip -2pt
\caption{Comparison of unlearning performance on the forget data of R-TOFU.}
\label{fig:unlearn_rtofu_forget}
\vskip -12pt
\end{figure*}

\begin{figure*}[t!]
\centering
\begin{subfigure}{0.495\linewidth}
\includegraphics[width=1\linewidth]{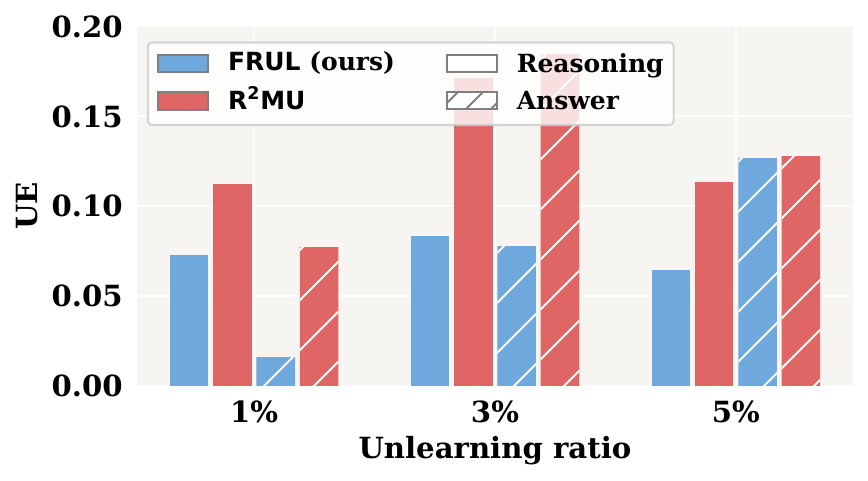}
\vskip -8pt
\caption{Reasoning-Llama-3.2-1B}
\end{subfigure}
\begin{subfigure}{0.495\linewidth}
\includegraphics[width=1\linewidth]{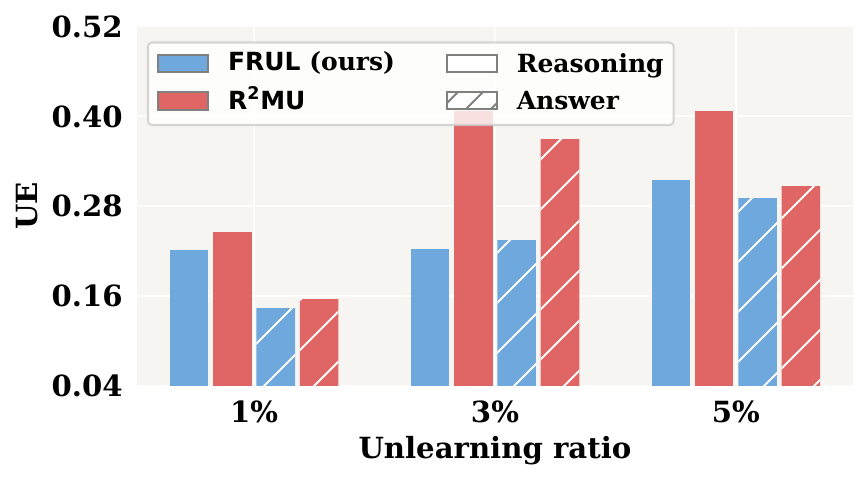}
\vskip -8pt
\caption{Nemotron-Nano-8B}
\end{subfigure}
\vskip -2pt
\caption{Comparison of retain utility on the retain data of R-TOFU.}
\label{fig:unlearn_rtofu_retain}
\end{figure*}

\textbf{Datasets.} We evaluate our proposed method on the following two datasets: R-TOFU~\cite{maini2024tofu}, a synthetic benchmark designed for controlled unlearning, and medical-o1-reasoning~\cite{chen2024huatuogpt}, a real-world medical reasoning corpus with rich domain-specific knowledge. The R-TOFU dataset contains 4,000 examples of question-answer pairs with detailed CoT reasoning, structured around fictional biographical information to enable controlled forgetting experiments \cite{yoon2025r,maini2024tofu}. R-TOFU provides a fictitious but semantically rich corpus that enables precise, reproducible comparisons between unlearned and original models. The medical-o1-reasoning dataset consists of 19,700 high-quality examples of chain of thought reasoning in complex clinical and biomedical contexts.

\begin{figure*}[t!]
\centering
\vskip -5pt
\begin{subfigure}{0.495\linewidth}
\includegraphics[width=1\linewidth]{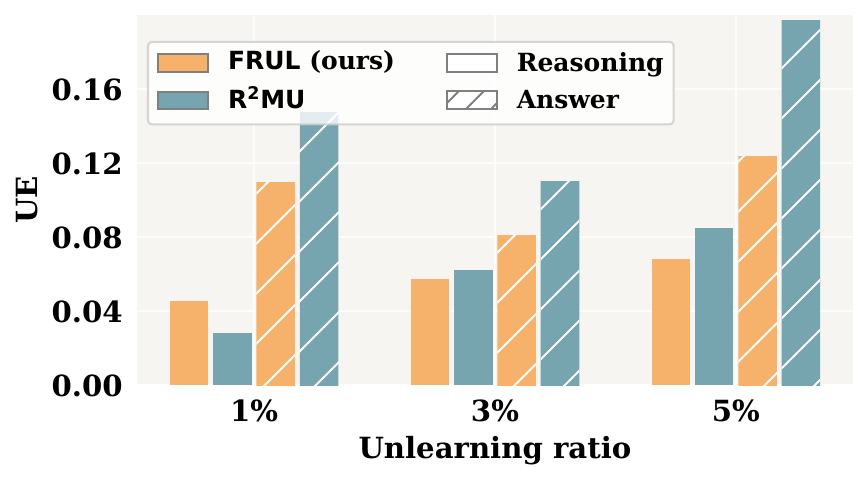}
\vskip -8pt
\caption{Reasoning-Llama-3.2-1B}
\end{subfigure}
\begin{subfigure}{0.495\linewidth}
\includegraphics[width=1\linewidth]{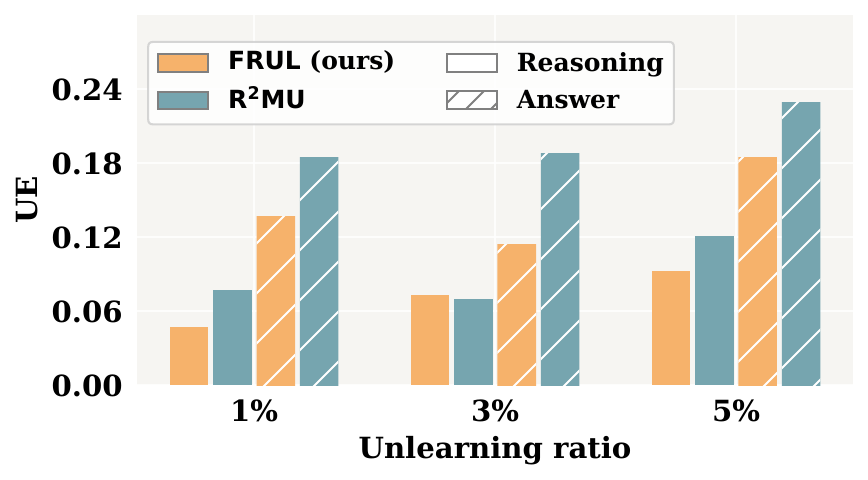}
\vskip -8pt
\caption{Nemotron-Nano-8B}
\end{subfigure}
\vskip -2pt
\caption{Comparison of unlearning performance on the forget data of the adopted medical-o1-reasoning dataset. }
\label{fig:unlearn_medqa_forget}
\vskip -12pt
\end{figure*}

\textbf{Models}. In experiments, we adopt the following LRMs: Reasoning-Llama-3.2-1B-Instruct-v1.2 from EpistemeAI
 and Llama-3.1-Nemotron-Nano-8B-v1 from Nvidia. Reasoning-Llama-3.2-1B is a compact 1.2B parameter model developed by EpistemeAI, designed specifically to support structured multi-step reasoning with minimal computational overhead. Its smaller size allows for fast experimentation while preserving key reasoning capabilities. Nemotron-Nano-8B from Nvidia is an 8B parameter model that provides stronger generalization and deeper reasoning capacity across complex prompts.

\textbf{Baselines}. We adopt \emph{Reasoning-aware Representation Misdirection Unlearning} (R$^2$MU)~\cite{wang2025rethinking} as our baseline. R$^2$MU performs unlearning by mapping the internal representations of both the entire CoT and the final answer of each sample in the forget set to a fixed random vector. To mitigate over-forgetting, it employs an auxiliary reasoning dataset to regularize the model and preserve general reasoning ability. We denote the unlearned model using R$^2$MU as $M_{R^{2}MU}$. We compare our method against R$^2$MU to show that our method achieves more efficient unlearning and superior reasoning preservation without requiring any external regularization dataset. We denote our unlearned method as $M_{FRUL}$.

\begin{figure*}[t!]
\centering
\begin{subfigure}{0.495\linewidth}
\includegraphics[width=1\linewidth]{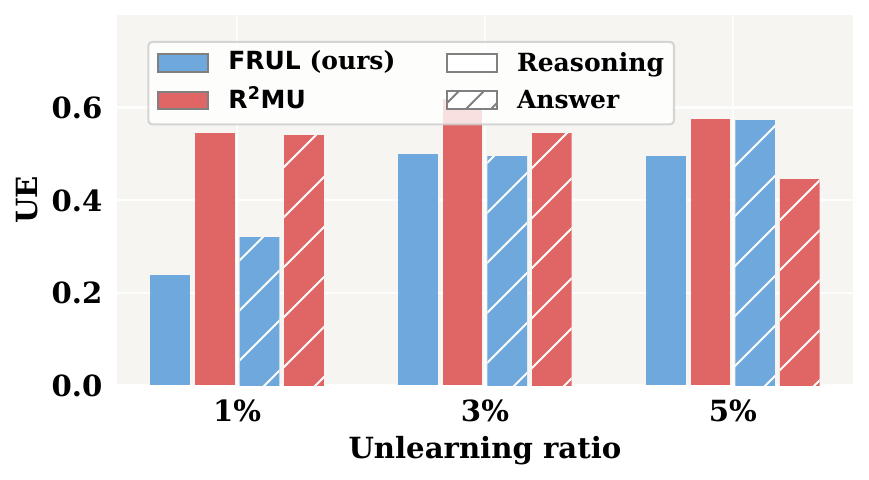}
\vskip -8pt
\caption{Reasoning-Llama-3.2-1B}
\end{subfigure}
\begin{subfigure}{0.495\linewidth}
\includegraphics[width=1\linewidth]{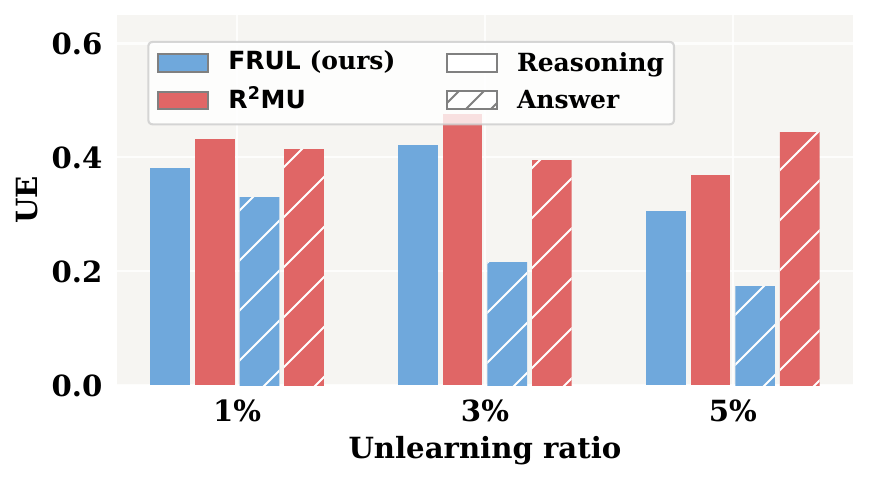}
\vskip -8pt
\caption{Nemotron-Nano-8B}
\end{subfigure}
\vskip -2pt
\caption{Comparison of retain utility on the retain data of medical-o1-reasoning.}
\label{fig:unelarn_medqa_retain}
\vskip -12pt
\end{figure*}

\textbf{Implementation Details.} In experiments, since the R-TOFU dataset’s content is synthetic and easily reconstructible, we can finetune the target LRM on $D_r$ to achieve exact unlearning conditions without dependency on proprietary pretraining data. In contrast, the medical dataset was not used in the pretraining or fine-tuning of our target model, this external dataset enables us to achieve exact unlearning when fine-tuning on $D_r$. To simulate unlearning, we randomly partition the dataset into forget and retain sets by selecting $1\%$, $3\%$, and $5\%$ of examples as $D_f$, with the remaining $99\%$, $97\%$, and $95\%$ as $D_r$, respectively. We denote the model fine-tuned only on $D_r$ as $M_r$, which serves as a target reference for successful unlearning. For each $(q, c, a) \in D_f$, we decompose $c$ into a forget-relevant segment $c_f$ and its benign replacement $c_m$ using multiple LLMs with RAG. This ensures that $c_m$ maintains semantic and logical fidelity to the original reasoning while omitting sensitive or undesired knowledge. For the trade-off parameters, we set \(\alpha=1\) in Eq.~(\ref{eq:loss_gd}),  \(\lambda_f=1\) and \(\lambda_r=2\) in Eq.~(\ref{eq:loss_cot}), \(\beta_g=0.25\) and \(\beta_r=0.75\) in Eq.~(\ref{eq:overall_loss}). These hyperparameter choices provide robust performance without requiring additional tuning. Note that to address unreliability in the extracted data information (primarily arising from LLM hallucinations), we adopt the uncertainty-aware weighted aggregation by using two distinct LLMs (i.e., \texttt{gpt-3.5-turbo} \cite{anand2023gpt4all} and \texttt{gpt-4.1-mini} \cite{openai2024gpt4}) to independently extract candidate segments, and then perform the aggregation of these results. We then use another LLM (\texttt{gpt-3.5-turbo}) to replace $c_f$ and generate $c_m$.

\textbf{Evaluation Metrics.} We evaluate our framework along two key dimensions: unlearning efficiency and general reasoning retention. Unlearning efficiency quantifies how effectively the unlearned model $M_{\text{FRUL}}$ removes knowledge associated with $D_f$ relative to the reference model $M_{r}$. General reasoning retention assesses the extent to which $M_{\text{FRUL}}$ preserves reasoning structures and outputs for the retain set $D_r$ compared to $M_r$. To measure these, we compute the divergence in reasoning responses and final answer between $M_{\text{FRUL}}$ and $M_{r}$ using the ROUGE metric~\cite{lin-2004-rouge}, which assesses word-level overlap between generated text and reference outputs. We define unlearning error (UE) as a metric to measure the difference between the ROUGE-L of generated response from $M_{\text{FRUL}}$ and $M_r$. A lower UE scores indicate closer alignment between $M_{\text{FRUL}}$ and $M_r$, and thus higher unlearning effectiveness. Formally, unlearning efficiency is defined as \(UE =|| \text{ROUGE}(M_{\text{FRUL}}) - \text{ROUGE}(M_{r})||\). We apply UE to the reasoning traces to quantify the model’s ability to preserve coherent and accurate reasoning while forgetting sensitive information in the CoTs after unlearning. We further apply the same metric to the final answers to demonstrate that our model achieves comparable or superior answer accuracy relative to R$^2$MU, thereby achieving effective unlearning without sacrificing end-task performance.

\begin{figure*}[t!]
    \centering
    \begin{subfigure}[b]{0.328\textwidth}
        \includegraphics[width=\textwidth]{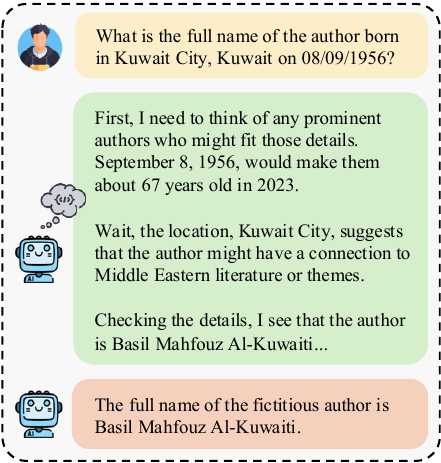}
        \vskip -2pt
        \caption{Original}
        \label{fig:original}
    \end{subfigure}
    \hfill
    \begin{subfigure}[b]{0.328\textwidth}
        \includegraphics[width=\textwidth]{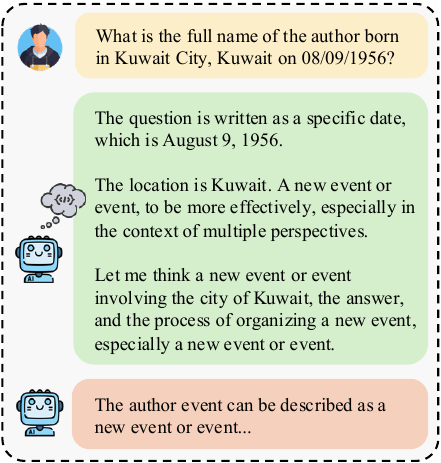}
        \vskip -2pt
        \caption{R$^2$MU}
        \label{fig:rmu}
    \end{subfigure}
    \hfill
    \begin{subfigure}[b]{0.328\textwidth}
        \includegraphics[width=\textwidth]{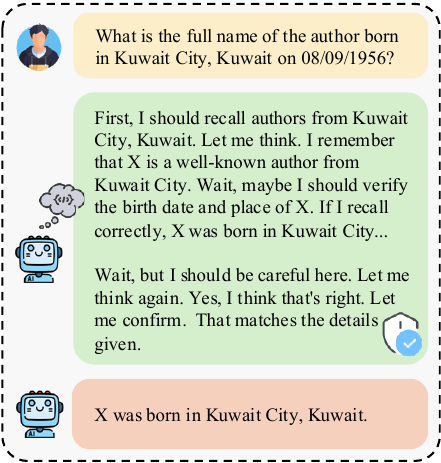}
        \vskip -2pt
        \caption{FRUL (ours)}
        \label{fig:frul}
    \end{subfigure}
\vskip -2pt
    \caption{Performance comparison of Reasoning-Llama-3.2-1B on the R-TOFU benchmark before and after unlearning on a 5$\%$ unlearning ratio. (a) Baseline model fine-tuned on the full R-TOFU training set; (b) Model unlearned using the R$^2$MU method; (c) Model unlearned using the proposed FRUL loss. }
    \label{fig:compare_unlearn}
    \vskip -12pt
\end{figure*}

\subsection{Main Results}
\vspace{-0.02 in}

We start by evaluating the effectiveness of FRUL. Specifically, we compare the performance gap in reasoning and final answers against the golden retraining approach using the UE metric, and adopt the R$^2$MU as the baseline. We evaluate the efficacy of the fine-tuning process by comparing ROUGE-L scores of reasoning and answer against the pre-trained baselines. The results demonstrate substantial performance gains across both datasets. For R-TOFU, Reasoning-Llama-3.2-1B's reasoning and answer scores improve from 0.0067 and 0.0133 to 0.4472 and 0.5932, respectively. For Medical-o1, scores improved from 0.1586 and 0.0244 to 0.5935 and 0.5943. Notably, the low ROUGE scores of the pretrained models confirm that they do not possess prior knowledge of either dataset, which is a desirable property for machine unlearning studies. Fig.~\ref{fig:unlearn_rtofu_forget} presents unlearning performance for reasoning and answers across different unlearning ratios on R-TOFU, using the Reasoning-Llama-3.2-1B and Nemotron-Nano-8B LRMs. The corresponding retain utilities for reasoning and answers after unlearning are shown in Fig.~\ref{fig:unlearn_rtofu_retain}. We first find that our proposed unlearning method achieves better reasoning performance in forgetting the sensitive information in reasoning traces. This is due to the fact that FRUL loss penalizes the generation of forgotten reasoning and encourages the model to replace it with non-sensitive content. Additionally, FRUL achieves comparable or even superior answer forgetting performance compared to the baseline. For example, for a 3\% unlearning ratio in Fig.~\ref{fig:unlearn_rtofu_forget_a}, FRUL yields an answer gap of 0.03 relative to retraining, whereas the baseline exhibits a gap of 0.06. The GD loss successfully eliminates answers for the forgotten questions. Furthermore, from the retained utility in Fig.~\ref{fig:unlearn_rtofu_retain}, we observe that our proposed unlearning method maintains strong reasoning and answer performance after unlearning. This is attributed to the design of the GD loss, which preserves answer utility, and the RP loss, which retains reasoning capability. Part of the CoT loss also helps the unlearned model to preserve the general reasoning ability.

\begin{figure*}[t!]
\centering
\begin{subfigure}{0.495\linewidth}
\includegraphics[width=1\linewidth]{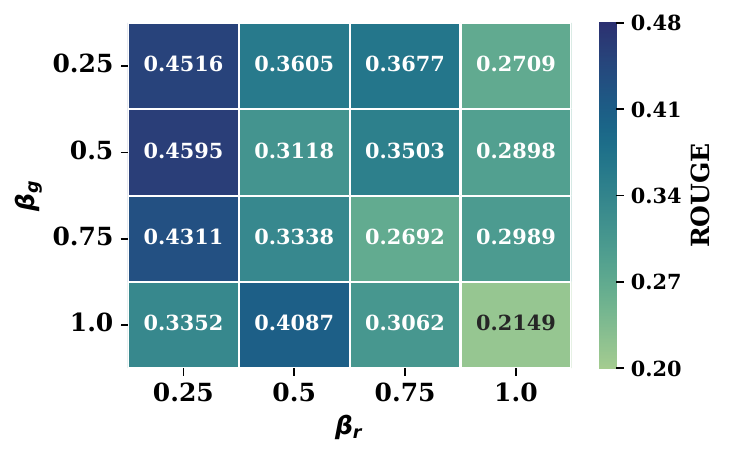}
\vskip -10pt
\caption{Answer unlearn performance}
\end{subfigure}
\begin{subfigure}{0.495\linewidth}
\includegraphics[width=1\linewidth]{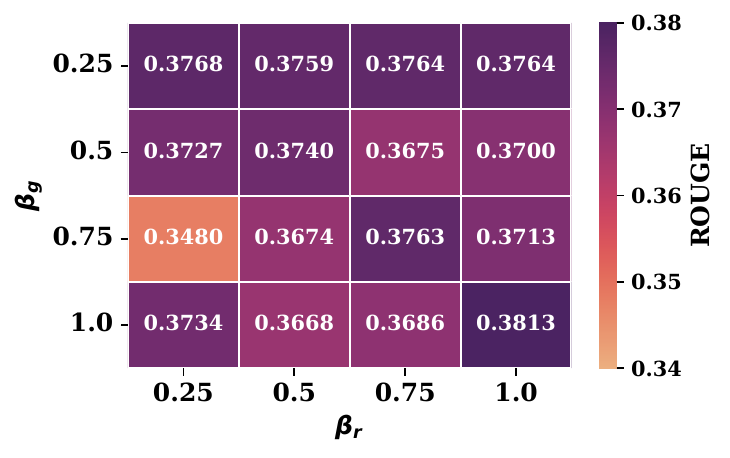}
\vskip -10pt
\caption{Reasoning retain performance}
\end{subfigure}
\vskip -2pt
\caption{Impact of trade-off hyperparameters on answer and reasoning.}
\label{fig:ablation_beta}
\vskip -12pt
\end{figure*}

Next, we validate the effectiveness of our proposed unlearning method on the medical-o1-reasoning dataset. As shown in Figs.~\ref{fig:unlearn_medqa_forget} and~\ref{fig:unelarn_medqa_retain}, FRUL achieves unlearning effectiveness on par with or better than R$^2$MU across various forget ratios, while also yielding lower UE scores on the retain set in both reasoning and final answers. In particular, Fig.~\ref{fig:unlearn_medqa_forget} illustrates that FRUL exhibits lower UE on $D_f$ in reasoning, attributed to the CoT loss, which encourages the model to maintain a coherent reasoning structure through benign replacements. Meanwhile, Fig.~\ref{fig:unelarn_medqa_retain} shows that our method has lower UE in both reasoning and the final answer, as the RP loss enhances performance in reasoning, and the gradient descent loss enhances the final answer in the $D_r$. This consolidates FRUL’s ability to precisely suppress forgotten content in reasoning and final answers without harming general reasoning ability on $D_r$, even in high-stakes domains like medical QA. Overall, our proposed unlearning method shows high effectiveness in forgetting answers and removing sensitive reasoning information, while preserving the model’s overall reasoning and answering ability.

Then, we provide visualization results of our proposed unlearning method in Fig.~\ref{fig:compare_unlearn}. Here, we adopt one sample in $D_f$ of the R-TOFU dataset. Fig.~\ref{fig:original} displays the response generated by $M_{\text{original}}$, which contains both the sensitive reasoning steps and the correct final answer. Figs.~\ref{fig:frul} and~\ref{fig:rmu} show the outputs of the unlearned models produced by FRUL and R$^2$MU, respectively. Both methods successfully remove the forgettable content and prevent answering the question. However, the response generated by R$^2$MU often exhibits ungrammatical or incoherent reasoning, lacking clear structure or meaning, as it maps the internal representation of the entire CoT to a random vector. In contrast, the output from FRUL preserves a coherent reasoning structure with sensitive content replaced by benign placeholders. This demonstrates that our method FRUL effectively suppresses forgotten knowledge while maintaining logical coherence and fluency in reasoning, attributed to the CoT-aware loss.

Furthermore, we perform ablation studies to examine the effectiveness of our proposed unlearning method under different trade-off hyperparameters associated with the GD loss and RP loss in Eq.~(\ref{eq:overall_loss}). Fig.~\ref{fig:ablation_beta} illustrates the impact of unlearning hyperparameter \(\beta_g\) and the reasoning preservation parameter \(\beta_r\) on answer unlearning performance and reasoning retain ability, evaluated on the R-TOFU dataset with Reasoning-Llama-3.2-1B. As shown, increasing \(\beta_g\) strengthens the gradient descent optimization on the forget set, improving unlearning efficiency for forgotten answers. However, a larger \(\beta_g\) may lead to over-unlearning, producing a larger gap from the retraining method and potentially degrading retain utility. In contrast, increasing \(\beta_r\), which promotes reasoning ability preservation on the retain data, steadily enhances the reasoning performance after unlearning. These findings underscore the importance of jointly tuning the trade-off hyperparameters to balance unlearning and retain performance in LRMs.

\section{Conclusion}
\label{sec:Conclusion}

In this work, we study how to achieve effective unlearning in large reasoning models without compromising their general reasoning performance. Towards this goal, we present a novel LRM unlearning framework that extends beyond final answers to selectively unlearn sensitive information embedded within reasoning steps. Specifically, in our proposed method, to precisely identify sensitive reasoning knowledge, we first leverage multiple LLMs with RAG to identify forget-relevant reasoning segments within the CoT. Then, we replace these segments with logically consistent and benign placeholders to preserve structural coherence. We further introduce a feature replacement-guided unlearning loss for LRMs, which jointly suppresses the generation of forgotten content and reinforces valid reasoning traces while incorporating reasoning preservation and gradient difference objectives to maintain performance on non-forgotten data. Extensive experiments have been made on both real-world and synthetic datasets and popular large reasoning models. We also provide insightful analysis for the effectiveness of our proposed method in erasing the targeted forget data in LRMs.

\subsubsection{\ackname} This work is supported in part by the US National Science Foundation under grants CNS-2350332 and IIS-2442750. Any opinions, findings, and conclusions or recommendations expressed in this material are those of the author(s) and do not necessarily reflect the views of the National Science Foundation.

%
%
\bibliographystyle{splncs04}
\bibliography{reference}

\end{document}